\pdfoutput=1
\documentclass[11pt]{article}

\usepackage{ACL2023}

\usepackage{times}
\usepackage{latexsym}

\usepackage{amsfonts}
\usepackage{tabularx}
\usepackage{booktabs}
\usepackage{multirow}
\usepackage{graphicx}
\usepackage{amsmath}
\usepackage{subfig}
\usepackage{diagbox}
\usepackage[T1]{fontenc}
\usepackage[utf8]{inputenc}

\usepackage{microtype}
\usepackage{inconsolata}

\title{DialoGPS: Dialogue Path Sampling in Continuous Semantic Space \\for Data Augmentation in Multi-Turn Conversations}

\author{Ang Lv$^{1}$\thanks{\ \ Equal contribution.}, Jinpeng Li$^{2}$\footnotemark[1], Yuhan Chen$^1$, Xing Gao$^{3}$, Ji Zhang$^{3}$, Rui Yan$^{1,4}$\thanks{\ \ Corresponding author: Rui Yan (\url{ruiyan@ruc.edu.cn}).} \\
  $^1$Gaoling School of Artifical Intelligence, Renmin University of China\\
  $^2$Wangxuan Institute of Computer Technology, Peking University\\
    $^3$Alibaba DAMO Academy\\
  $^4$Engineering Research Center of Next-Generation Intelligent \\ Search and Recommendation, Ministry of Education\\
  \texttt{\{anglv, yhchen, ruiyan\}@ruc.edu.cn, lijinpeng@stu.pku.edu.cn, } \\
  \texttt{\{gaoxing.gx,zj122146\}@alibaba-inc.com} \\
}

\begin{document}
\maketitle
\begin{abstract}
In open-domain dialogue generation tasks, contexts and responses in most datasets are one-to-one mapped, violating an important many-to-many characteristic: a context leads to various responses, and a response answers multiple contexts. Without such patterns, models poorly generalize and prefer responding safely. Many attempts have been made in either multi-turn settings from a one-to-many perspective or in a many-to-many perspective but limited to single-turn settings. The major challenge to many-to-many augment multi-turn dialogues is that discretely replacing each turn with semantic similarity breaks fragile context coherence. In this paper, we propose DialoGue Path Sampling (DialoGPS) method in continuous semantic space, the first many-to-many augmentation method for multi-turn dialogues. Specifically, we map a dialogue to our extended Brownian Bridge, a special Gaussian process. We sample latent variables to form coherent dialogue paths in the continuous space. A dialogue path corresponds to a new multi-turn dialogue and is used as augmented training data. We show the effect of DialoGPS with both automatic and human evaluation.
\end{abstract}

\section{Introduction}
\label{sec:intro}

Open-domain dialogue generation has received significant attention and has made notable advancements~\cite{zhang2019dialogpt,blendorbot,chatgpt}. However, it still faces challenges due to the nature of the data. One specific challenge is the many-to-many relationship between contexts and responses in open-domain conversations. A context can lead to various responses, and a response can be relevant to multiple contexts. Unfortunately, most datasets only provide one-to-one mappings between contexts and responses. This limitation results in models being poorly generalized when they rely on learned one-to-one patterns, making them prone to generating safe yet uninteresting responses~\cite{jiang-de-rijke-2018-sequence,10.1145/3308558.3313415}.

\begin{figure}[t]
\centering
\subfloat[Discrete replacement causes incoherence.]{\includegraphics[width=0.85\linewidth]{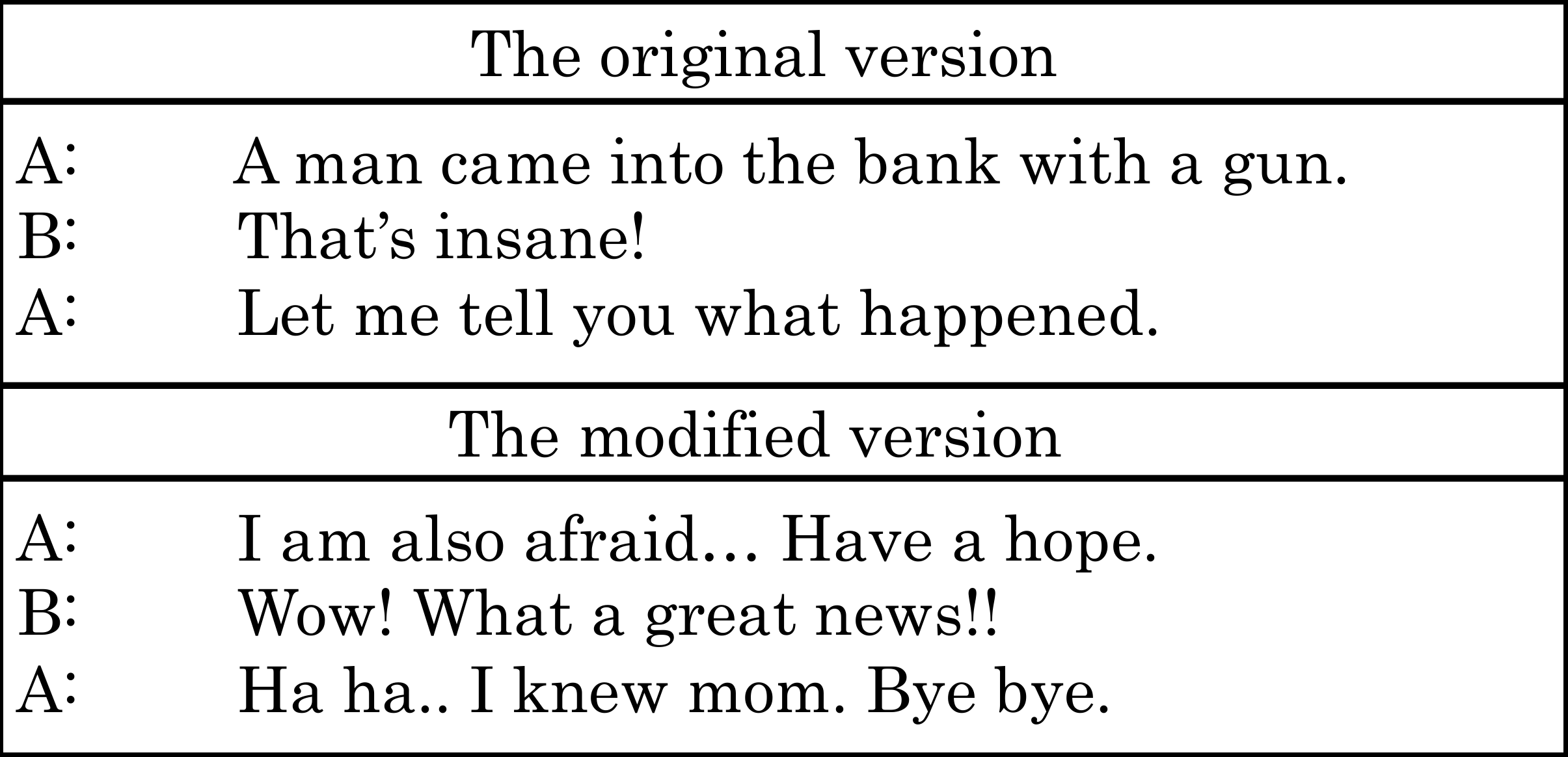}}

\subfloat[Sampled dialogue paths in the continuous semantic space correspond to coherent discrete dialogues.]{\includegraphics[width=\linewidth]{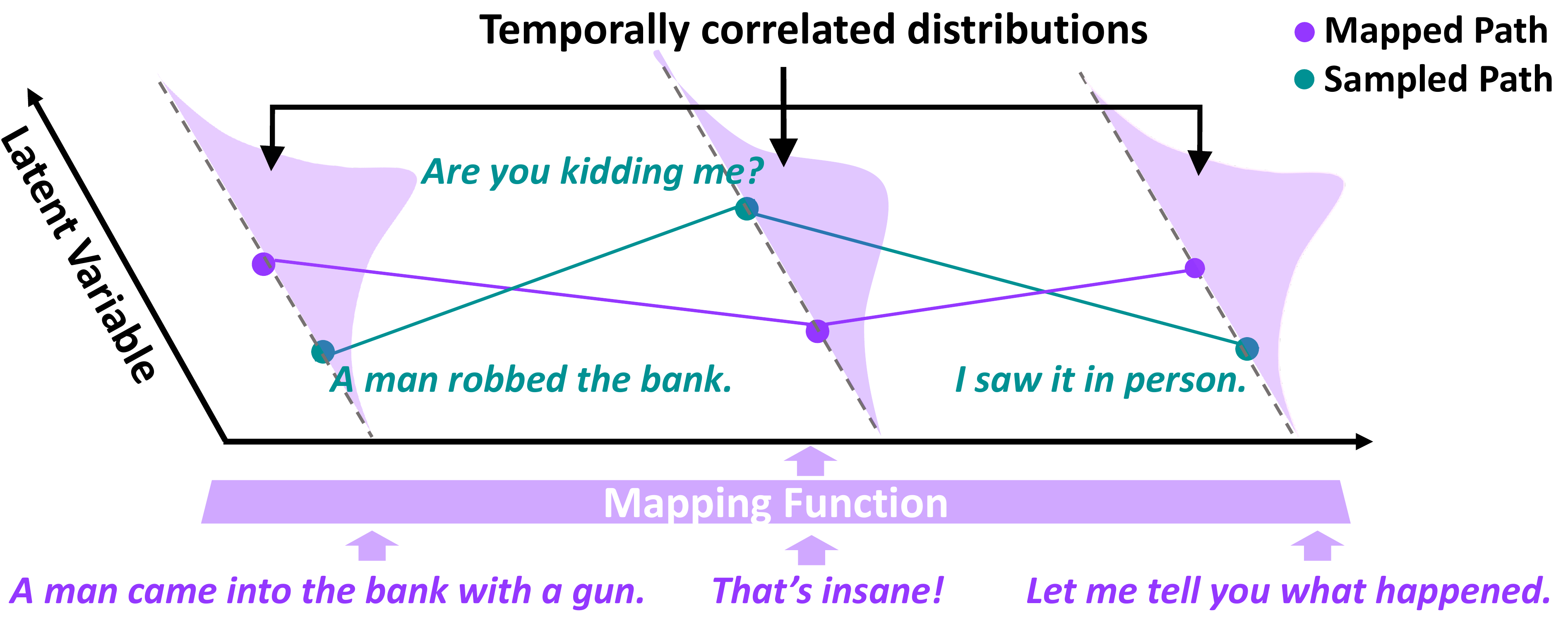}}
\caption{(a) When replacing each utterance in the original conversation by semantic similarity, the modified dialogue is incoherent. (b) We map dialogues into a continuous semantic space where latent distributions of utterances correlate with each other, and sample dialogue paths for training. Each path corresponds to a discrete multi-turn conversation.}
\label{fig:motivation}
\end{figure}

To address this limitation, many attempts~\cite{sai-etal-2020-improving,qiu-etal-2019-training,xie2022targetside} have been made from a one-to-many perspective which involves constructing multiple responses for a context. 
Furthermore, some works are proposed from a many-to-many perspective but are limited to single-turn settings. 
To construct new dialogue sentence pairs, they either replace sentences based on semantic similarity~\cite{zhang-etal-2020-dialogue} or sample new sentences from probabilistic models~\cite{Li_Qiu_Tang_Chen_Zhao_Yan_2019}.
Next, they adopt BERT~\cite{devlin-etal-2019-bert} or GAN~\cite{NIPS2014_5ca3e9b1} discriminators to filter incoherent sentence pairs. 

These methods cannot be trivially extended to multi-turn settings. Considering $T$ utterances in a dialogue and $K$ candidates for each utterance, they need to (1) prepare a large sentence set as candidates for replacement or a strong generative model, and (2) check the coherence of the modified conversation at least $K^{T-1}$ times, which is impractical.
Figure~\ref{fig:motivation}(a) shows a case in which we replace each utterance in a conversation following~\citet{zhang-etal-2020-dialogue}. 
The modified conversation is still incoherent across turns. 
Therefore, to enhance multi-turn dialogue generation from a many-to-many perspective, we resort to a continuous semantic space that satisfies two requirements. 
First, it describes semantic distributions of utterances, allowing for sampling semantic neighbors of each utterance. 
Second, latent variables sampled from any two distributions should be temporally correlated, contributing to a new coherent dialogue path in the latent space without requiring post-checks. 
This path can be utilized as a new training sample to augment the model. 
Our motivation is illustrated in Figure~\ref{fig:motivation}(b).

Driven by this motivation, we propose a novel method for augmenting open-domain dialogues from a many-to-many perspective, called Dialo\textbf{G}ue \textbf{P}ath \textbf{S}ampling (DialoGPS), aiming to enhance generalization and improve the quality of generated responses.
Specifically, our approach involves the following steps:
(1) We map each utterance in a multi-turn dialogue to a special Gaussian process in a continuous semantic space known as the Brownian Bridge~\cite{revuz2013continuous}.
(2) For each utterance $x_i$, we sample $K$ latent variables $z^{j}_i$, $j \in \left[1,K\right]$, establishing $K$ different dialogue paths in the bridge. Each path corresponds to a new multi-turn conversation in the discrete space.
(3) DialoGPS utilizes an encoder-decoder architecture. To construct augmented data, we mix the latent variable $z_i$ with representations of $x_i$ in the encoder if $x_i$ is part of the context, and in the decoder if it is the response.
(4) Finally, we train the model using the augmented data.

To ensure the effectiveness of DialoGPS, we address several key issues.
First, traditional Brownian Bridges have deterministic endpoints, which prevent response sampling and lead our method degenerating into a many-to-one paradigm, further impairing generalization. To overcome this limitation, we derive the formula of endpoint distributions. 
Second, since augmented data that lacks discrete utterance labels makes the optimization challenging, we propose a self-distillation framework where the model first learns from the ground truth and then distills its knowledge to guide itself in utilizing augmented data.

We evaluate DialoGPS on two multi-turn open-domain datasets. Both automatic and human evaluation show that DialoGPS performs better than strong baselines and even outperforms the model trained on manually denoted multi-reference data, which demonstrates the benefit of the many-to-many augmentation paradigm. Because DialoGPS is plug-and-play, we add it to BART~\cite{lewis-etal-2020-bart} and achieve competitive results with the state-of-the-art model, DialoFlow~\cite{li-etal-2021-conversations}. Our contributions are as follows: 

$\bullet$ DialoGPS is the first work to augment multi-turn dialogues from a many-to-many perspective.

$\bullet$ To ensure the effectiveness of DialoGPS, we have introduced dialogue-specific designs, including endpoint sampling of Brownian Bridges and self-distillation for model optimization.

$\bullet$ Experiments conducted on both non-pretrained and pre-trained models show that our DialoGPS method outperforms all baselines.

\section{Related Work: Dialogue Generation Augmentation}
% \subsection{Dialogue Generation}
In general, dialogue generation can be categorized into two groups: task-oriented and open-domain. Open-domain generation is a context-aware process that lasts for turns. The model learns to generate a proper but open response from the preceding utterances (i.e., contexts). Task-oriented dialogues progress for specific purposes and are limited to specific domains, such as obtaining knowledge~\citep{zhao-etal-2020-knowledge-grounded,tao2021pre}. 
However, due to the specific domains in task-oriented dialogues, the many-to-many relationship is not as apparent compared to open-domain dialogues.

In this paper, we focus on open-domain dialogue generation augmentation from an $X$-to-many perspective. From a one-to-many perspective, \citet{sai-etal-2020-improving} manually denoted multiple responses for a dialogue context. Based on such multi-reference datasets, \citet{qiu-etal-2019-training} proposed to capture the common feature in feasible responses and then add the specific feature to obtain the final output, which augments the utility of the data and improves the generalization. \citet{xie2022targetside} proposed that with only one-to-one data, models can construct pseudo-target data in the decoder and improve the model by bootstrapping. From a many-to-many perspective, existing methods work in single-turn settings. \citet{Li_Qiu_Tang_Chen_Zhao_Yan_2019} generated multiple context or responses with CVAE~\cite{zhao-etal-2017-learning} and introduced a GAN~\cite{NIPS2014_5ca3e9b1} discriminator to filter incoherent sentence pairs. \citet{zhang-etal-2020-dialogue} augmented a one-to-one dialogue dataset $D_p$ with an unpaired sentence set $D_u$. They sample sentences from $D_u$ and replace the most similar sentences in $D_p$. They use BERT~\cite{devlin-etal-2019-bert} and knowledge distillation to filter noise in incoherent sentence pairs.
Until now, many-to-many augmentation in multi-turn settings are understudied.

\begin{figure*}[!t]
\subfloat[Method overview.]{\includegraphics[scale=0.19]{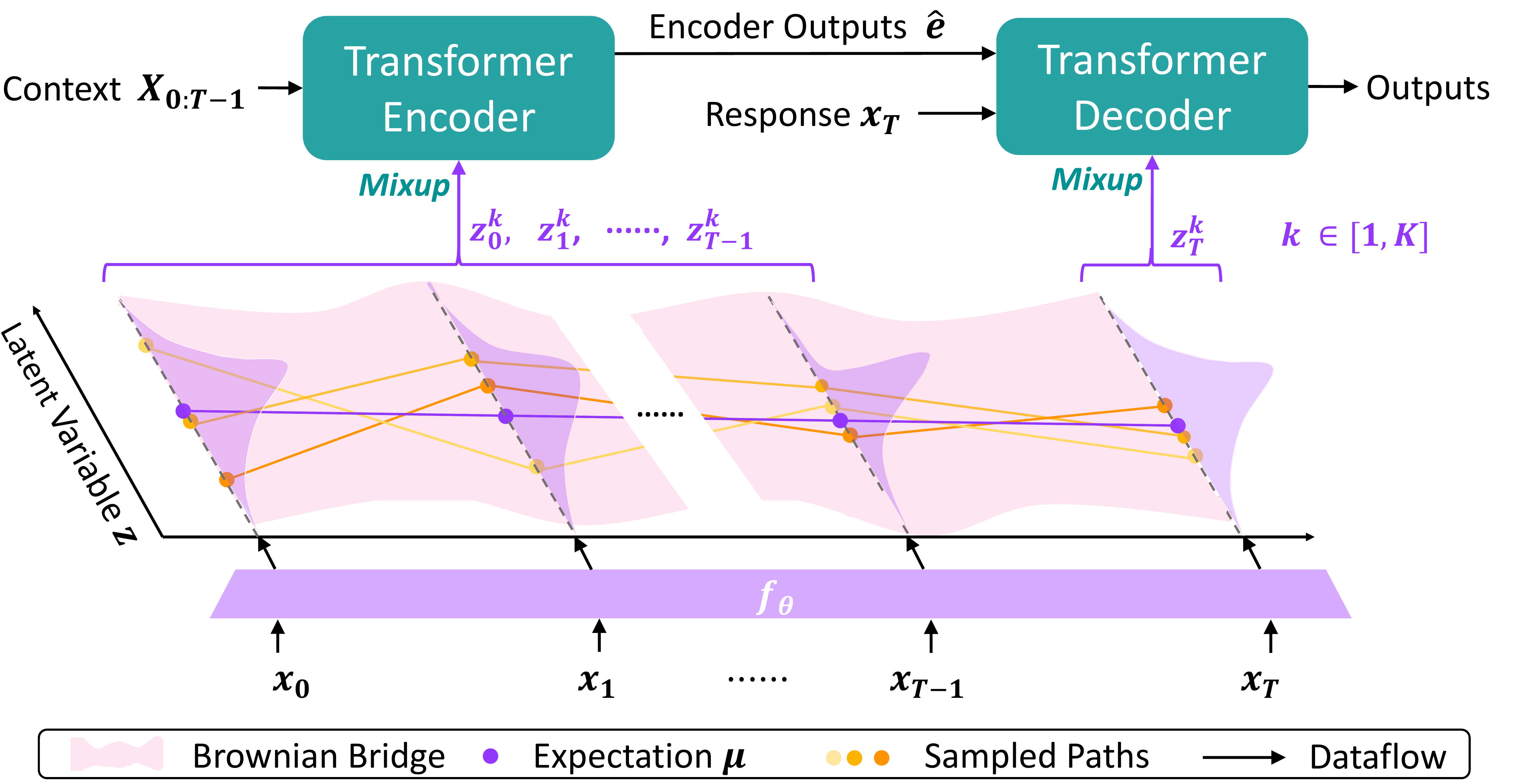}}
\subfloat[Mixup details on encoder and decoder.]{\includegraphics[scale=0.13]{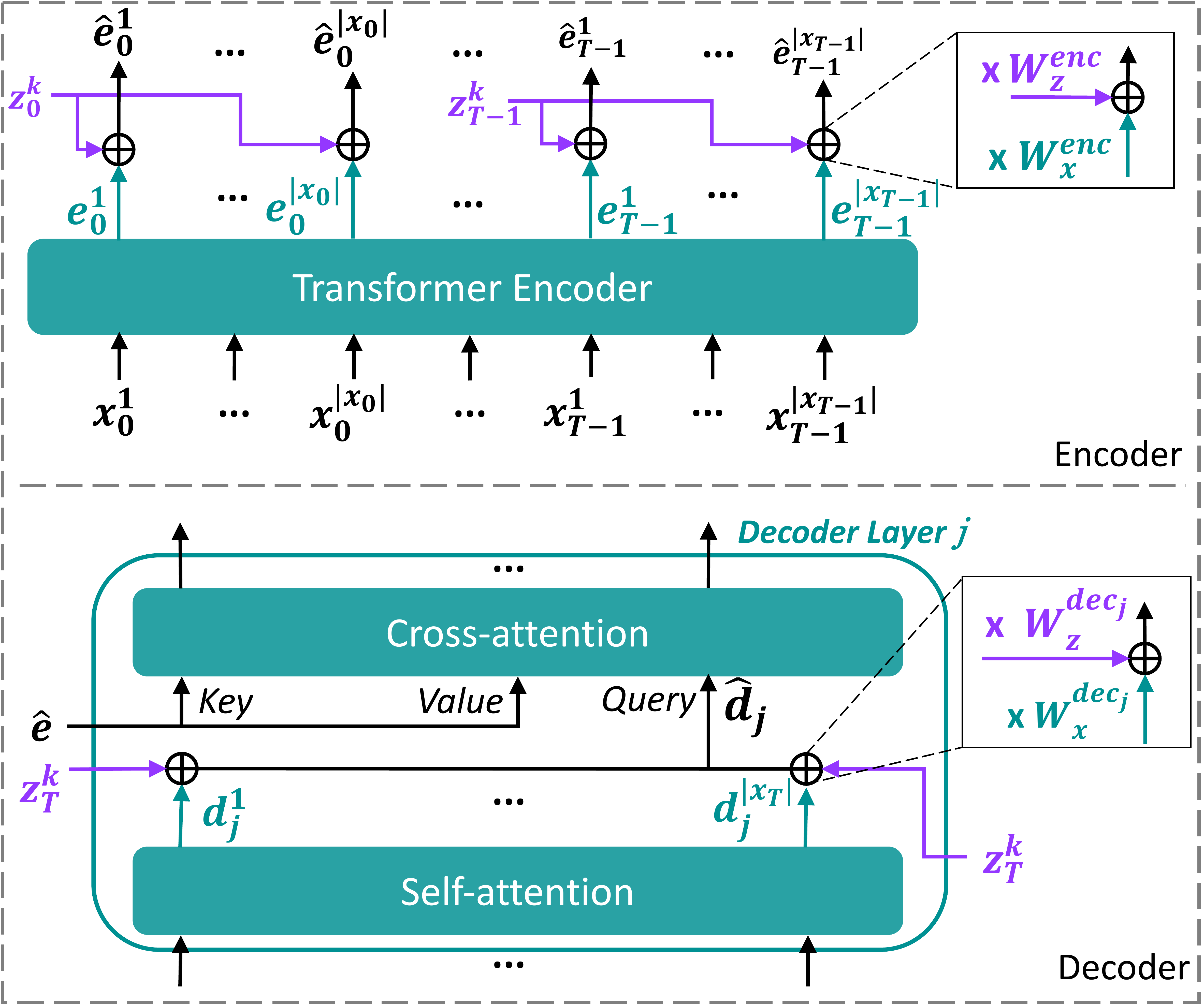}}
\caption{(a) The overview of DialoGPS. Teacher forcing is applied during training. Each utterance in the dialogue is mapped into a semantic distribution on a Brownian Bridge. We sample $K$ paths and conduct mixup operations in the encoder and decoder, respectively. (b) Mixup details.}
\label{fig:model}
\end{figure*}

\section{Method}

We first present some preliminaries ($\S$~\ref{sec:preliminary}). Then, we introduce mapping dialogue texts to the desired latent space ($\S$~\ref{sec:my-brownian-bridge}), augmented data construction ($\S$~\ref{sec:mixup}), augmented data utilization ($\S$~\ref{sec:self-distill}), and inference details ($\S$~\ref{sec:infer}). Figure \ref{fig:model} shows the overview of DialoGPS.

\subsection{Preliminary}
\label{sec:preliminary}

In open-domain dialogue generation, given a multi-turn dialogue $X = \left[x_0, x_1, ..., x_T\right]$, the goal is to predict the response $x_T$ based on the context $X_{0:T-1}$. The number of tokens in $x_t$ is denoted as $|x_t|$, $t \in \{0, 1, \dots, T\}$. The $i$-th token in the $x_t$ is denoted as $x^{i}_t$. A Brownian Bridge $\mathcal{B}$ defined on time range $[0,T]$ is a special Gaussian process established on deterministic endpoints $\mu_0$ and $\mu_T$. At time $t$, the latent variable $z_t$ follows a Gaussian distribution $\mathcal{B}(t|\mu_0,\mu_T)$: 
\begin{equation}
\small
    z_t \sim \mathcal{B}(t|\mu_0,\mu_T) = \mathcal{N}(\mu_0 + \frac{t}{T}(\mu_T - \mu_0), \frac{t(T-t)}{T}),
\label{eq:brownian-defination}
\end{equation}

\subsection{Extended Brownian Bridge}
\label{sec:my-brownian-bridge}
In DialoGPS, given $X$, a non-linear function $f_{\theta}$ maps each $x_t$ to $\mu_t$, the expectations of the corresponding semantic distribution. Based on $\mu_0$ and $\mu_T$, we can establish a Brownian Bridge, and from which we sample the latent variable $z_{t}$ as the semantic neighbor of $x_t$. Meanwhile, $z_0, z_1, ..., z_T$ compose a coherent dialogue path because in a Brownian Bridge, the covariance between $t_1$ and $t_2$, with 0 < $t_1$ < $t_2$ < $T$ is $\frac{t_{1}(T-t_{2})}{T}$, where the constant positive covariance guarantees that $\mathcal{B}(t_{1}|\mu_0,\mu_T)$ and $\mathcal{B}(t_{2}|\mu_0,\mu_T)$ are temporally correlated. 

However, as defined in Eq.~\ref{eq:brownian-defination}, a conventional Brownian Bridge $\mathcal{B}$ has deterministic endpoints, which prevents us from sampling for $x_T$, the response, and $x_0$, the first utterance in the context. To avoid degenerating to a many-to-one mode that impairs the generalization, we derive an extended Brownian Bridge $\beta$ with samplable endpoints. Take the derivation of $\beta(T|\mu_0,\mu_T)$ as example: given a $\mathcal{B}$, both the distance $d_{\delta}$ between $\mu_T$ and $z_{T-\delta}$ and the summation of $d_{\delta}$ and $z_{T-\delta}$ follow the Gaussian distribution, we can derive the distribution of $z_T$ as follows:

\begin{equation}
\small
\begin{aligned}
    \left.
    \begin{aligned}
    z_{T-\delta} \sim  \mathcal{N}(\frac{T - \delta}{T}\mu_T + \frac{\delta}{T}\mu_0, \frac{\delta(T-\delta)}{T})\\
    d_{\delta} = \mu_{T} - z_{T-\delta} \sim \mathcal{N}(\frac{\delta}{T}\mu_T - \frac{\delta}{T}\mu_0, \frac{\delta(T-\delta)}{T}) \\
    \end{aligned}
    \right\} \Rightarrow
     & \\ z_T = d_{\delta} + z_{T-\delta} \sim \mathcal{N}(\mu_T, \frac{2\delta(T-\delta)}{T}). \ \ \ \ \ \ \ \ \ \ \ \ 
\end{aligned}
\end{equation}

Due to the symmetry, $z_0$ follows $\mathcal{N}(\mu_0, \frac{2\delta(T-\delta)}{T})$. Here, $\delta$ serves as a hyper-parameter. To sum up, we define the extended Brownian Bridge $\beta$ as:
\begin{equation}
\small
\resizebox{\linewidth}{!}{
$\begin{aligned}
    \beta(t|\mu_0,\mu_T)=\left\{
    \begin{aligned}
    &\mathcal{N}(\mu_t, \frac{2\delta(T-\delta)}{T}) \mbox{, t = 0 or T}, \\
     &\mathcal{N}(\mu_0 + \frac{t}{T}(\mu_T - \mu_0), \frac{t(T-t)}{T}), \mbox{ otherwise}.\\
    \end{aligned}
    \right. 
\end{aligned}$}
\end{equation}

To optimize the mapping function $f_\theta$, we follow~\cite{wang2022language} to adopt a contrastive learning framework where positive samples are ordered sentence triplets from the same conversation ($x_{t_0}$, $x_{t_1}$, $x_{t_2}$, $t_0 < t_1 < t_2$) and negative samples are constructed by randomly replacing the middle point $x_{t_1}$ with other sentences $x_{t^{'}_1}$ from the mini-batch $\mathbb{B}$. The objective is as below:
\begin{equation}
\small
\resizebox{\linewidth}{!}{
$\begin{aligned}
    &\mathcal{L}_{\beta} = \mathbb{E}_{X}\left[\log\left(1+\frac{\sum\limits_{(x_{t_0}, x_{t^{'}_1}, x_{t_2}) \in \mathbb{B}} \exp(d(x_{t_0}, x_{t^{'}_1}, x_{t_2}; f_\theta))}{\exp(d(x_{t_0}, x_{t_1}, x_{t_2}; f_\theta))}\right)\right],
\end{aligned}$}
\label{eq:brownian-loss}
\end{equation}
where $d(x_{t_0}, x_{t_1}, x_{t_2}; f_\theta) = -\frac{1}{2\sigma^{2}_{t_1}} \Vert f_{\theta}(x_{t_1}) - (1 - \frac{t_1}{t_2})f_{\theta}(x_{t_0}) - \frac{t_1}{t_2}f_{\theta}(x_{t_2}) \Vert^{2}_2$ .
The essence of Eq. \ref{eq:brownian-loss} is to optimize the outputs of $f_\theta$, i.e., $\mu_{t_0}$, $\mu_{t_1}$, and $\mu_{t_2}$ to the linear relationship as defined in Eq. \ref{eq:brownian-defination}. In DialoGPS, a 4-layer MLP serves as $f_\theta$. To embed utterance as inputs of $f_\theta$, there are many choices such as averaging token embeddings or encoding by a language model. We leave the embedding details in $\S$\ref{sec:utterance-representation}.

\subsection{Augmented Data Construction}
\label{sec:mixup}
As shown in Figure \ref{fig:model}(a), we take Transformer~\cite{vaswani2017attention} as the bone architecture. With $f_\theta$, an extended Brownian Bridge $\beta$ is established. We sample latent variables $z_{t} \sim \beta(t| \mu_0, \mu_T)$ and mix them with representations of corresponding $x_t$. In the encoder, for each utterance $x_t$ in the context $X_{0:T-1}$, we conduct:
\begin{equation}
\begin{aligned}
    &e^{1}_{t}, e^{2}_{t}, ... e^{|x_t|}_{t} = \text{Encoder}(x_t),\\
    &\hat{e}^{i}_{t} = W^{enc}_x \cdot e^{i}_{t} + W^{enc}_z \cdot z_{t}, \\
\end{aligned}
\end{equation}
\noindent where $e^{i}_{t}$ is the output corresponding to the $i$-th token in $x_t$ from the encoder, $i \in \left[1, |x_t|\right]$. $W^{enc}_z$ and $W^{enc}_x$ are trainable vectors of the same dimension as $e$ and $z$. Finally, $\hat{e}$ is sent to the decoder for cross-attention. We conduct the mixup every decoder layer:
\begin{equation}
\begin{aligned}
    &\hat{d}^{i}_{j} = W^{dec_j}_{x} \cdot d^{i}_{j} + W^{dec_j}_{z} \cdot z_T, \\ 
    & i \in \left[1, |x_T|\right], j \in \left[1,N\right],
    \label{eq:mixup-in-dec}
\end{aligned}
\end{equation}
\noindent where $N$ is the number of decoder layers, $d^{i}_{j}$ is the self-attention output at position $i$ in layer $j$. Also, $W^{dec_j}_{z}$ and $W^{dec_j}_{x}$ are trainable vectors. $\hat{d}_j$ is used as \textit{Query}, and $\hat{e}$ are used as both \textit{Key} and \textit{Value} in the cross-attention. For a dialogue text $X$, we conduct sampling and mixup $K$ times, which is equivalent to providing $K$ extra discrete dialogues $\hat{X}^{k} = \left[\hat{x}^{k}_{0}, \hat{x}^{k}_{1}, ..., \hat{x}^{k}_{T}\right]$, $k\in\left[1,K\right]$ for training. Figure \ref{fig:model}(b) shows mixup details.

\subsection{Utilizing Augmented Data by Self-Distillation}
\label{sec:self-distill}

In general, given $X$ to a dialogue generation model, parameters $\phi$ of model are optimized by minimizing the negative log-likelihood:
\begin{equation}
\small
   \phi = {\rm argmin} \left(\mathbb{E}_{X}\left[-\log(P_{\phi}(x_T|X_{0:T-1]}))\right] \right).
\label{eq:1to1-obj}
\end{equation}

However, as aforementioned, what we obtain are continuous representations of $\hat{X}$ whereas the corresponding discrete sentences are inaccessible, which makes Eq.~\ref{eq:1to1-obj} intractable. Hence, to utilize the augmented data, we make an assumption that: There is an inaccessible many-to-many dialogue dataset $D_{MtoM}$. $P_{MtoM}$ describes the conditional distribution of responses given contexts in this dataset. The accessible one-to-one dataset $D_{1to1}$ is collected by sampling from $D_{MtoM}$ uniformly, and thus $P_{1to1}$ can be viewed as an approximation of $P_{MtoM}$.

Based on this assumption, we propose a self-distillation framework consisting of two steps: (1) It optimizes the model with the original discrete data following Eq.~\ref{eq:1to1-obj}.
(2) During training, as $P_{\phi}$ fits $P_{1to1}$, which is an approximation of $P_{MtoM}$, the model can use its output given $X$ to teach itself when presented with augmented data, i.e., the representations of $\hat{X}$:
\begin{equation}
\small
\resizebox{\linewidth}{!}{
$\begin{aligned}
   \phi = {\rm argmin} \left(D_{KL}\left[P_{\phi}(x_{T}|X_{0:T-1}) || P_{\phi}(\hat{x}_{T}|\hat{X}_{0:T-1})\right] \right),
\end{aligned}$}
\label{eq:kl}
\end{equation}
where $D_{KL}[\cdot||\cdot]$ is the KL-divergence~\citep{kullback1951information}. In Eq.~\ref{eq:kl}, to remove the gap between utilizing the original discrete data $X$ and the augmented continuous data $\hat{X}$ in the same architecture, we mix each utterance in $X$ with the expectations $\mu_{0:T}$. Formally, the overall training objective is to minimize:

\begin{equation}
\small
\begin{aligned}
   &\mathcal{L} = \underbrace{\mathcal{L}_\beta}_{\text{Mapping $X$ to $\beta$}}+\quad\underbrace{\mathbb{E}_{X}\left[-\log(P_{\phi}(x_{T}|X_{0:T-1},\mu_{0:T}))\right]}_{\text{Utilizing original discrete data}}\quad+ \\
   & \underbrace{\frac{1}{K}\sum\limits^{K}_{k} D_{KL}\left[P_{\phi}(x_{T}|X_{0:T-1},\mu_{0:T}) || P_{\phi}(\hat{x}^{k}_T|\hat{X}^{k}_{0:T-1},z^{k}_{0:T})\right]}_{\text{Utilizing augmented data}}
\end{aligned}
\label{eq:obj}
\end{equation}

\subsection{Inference}
\label{sec:infer}
The inference goal is to predict $x_T$ based on context $X_{0:T-1}$.
First, $f_{\theta}$ takes $X_{0:T-1}$ and outputs corresponding $\mu_t$ for sampling and mixup in the encoder, where $t\in\{0,1,\dots,T-1\}$. 
Next, the decoder receives the encoder output and an inferred $\mu_T$ to decode the response in an autoregressive manner.
To obtain the value of $\mu_T$, we do not require additional prediction networks. Instead, we can directly derive its value based on the property of Brownian Bridge.
Specifically, given the context, we know that for any $t$:

\begin{equation}
\small
\begin{aligned}
    \mu_t = \mu_0 + \frac{t}{T-1}(\mu_{T-1} - \mu_0).
\end{aligned} 
\label{eq:infer}
\end{equation}

If $\mu_T$ is already known, a Brownian bridge established on $\mu_T$ and $\mu_0$ would yield the same $\mu_t$ values. Consequently, we can establish an equality and derive the value of $\mu_T$ as follows:
\begin{equation}
\small
\begin{aligned}
    &\mu_t = \mu_0 + \frac{t}{T}(\mu_{T} - \mu_0) = \mu_0 + \frac{t}{T-1}(\mu_{T-1} - \mu_0)\\
    &\Rightarrow \mu_{T} = \frac{T}{T-1}\mu_{T-1} - \frac{1}{T-1}\mu_0.
\end{aligned} 
\label{eq:inferred-z}
\end{equation}

We find that there is hardly a difference in evaluation results when conducting mixup operations with either expectations $\mu$ or sampled variables $z$. To reduce randomness for easier analyses, experiments in below use expectations $\mu$ to mixup. Nonetheless, sampling variables gives DialoGPS the ability to generate diverse responses to an arbitrary context and we will discuss it in $\S$~\ref{sec:sample-infer}.

\begin{table*}[t]
    \centering
    \vskip 0.147in
    \resizebox{0.95\linewidth}{75mm}{
    \begin{tabular}{@{}lccccccc@{}}
    \toprule
\textbf{Models}
      & \textbf{BLEU-1} & \textbf{BLEU-2}& \textbf{BLEU-3} & \textbf{BLEU-4}   & \textbf{DIST-1} & \textbf{DIST-2.} & \textbf{BLEURT} \\ \midrule
      \multicolumn{8}{c}{ PersonaChat Dataset}\\\midrule
    
    Transformer & $ 17.79_{\left[0.14\right]}$ & $ 6.93_{\left[0.06\right]}$ & $ 3.03_{\left[0.08\right]}$ & $ 1.41_{\left[0.06\right]}$  & $ 0.82_{\left[0.01\right]}$ & $ 6.60_{\left[0.05\right]}$  & $ 30.16_{\left[0.05\right]}$ \\
    ResBag & $ 17.82_{\left[0.17\right]}$ & $ 6.88_{\left[0.12\right]}$ & $ 3.04_{\left[0.09\right]}$ & $ 1.37_{\left[0.11\right]}$ & $ 0.85_{\left[0.02\right]}$ & $ 6.83_{\left[0.02\right]}$ & $ 30.25_{\left[0.17\right]}$\\
    TSA  & $ 17.76_{\left[0.19\right]}$ & $ 6.92_{\left[0.16\right]}$ & $ 2.97_{\left[0.15\right]}$ & $ 1.35_{\left[0.10\right]}$  & $ 0.85_{\left[0.02\right]}$ & $ 6.56_{\left[0.01\right]}$  & $ 30.66_{\left[0.09\right]}$\\
    M\&D-D & $ 18.42_{\left[0.13\right]}$ & $ 7.25_{\left[0.09\right]}$ & $ 3.23_{\left[0.11\right]}$ & $ 1.44_{\left[0.07\right]}$  & $ 0.80_{\left[0.01\right]}$ & $ 6.55_{\left[0.01\right]}$ & $ 30.46_{\left[0.13\right]}$\\\midrule
    $\text{DialoGPS}_{K=1}$ & $ 18.29_{\left[0.08\right]}$ & $ 7.21_{\left[0.05\right]}$ & $ 3.14_{\left[0.03\right]}$ & $ 1.44_{\left[0.05\right]}$  & $ \textbf{1.05}_{\left[0.01\right]}$ & $ \textbf{7.97}_{\left[0.07\right]}$ & $ 30.54_{\left[0.06\right]}$\\
    $\text{DialoGPS}_{K=2}$ & $ 18.96_{\left[0.15\right]}$ & $ 7.61_{\left[0.09\right]}$ & $ 3.32_{\left[0.04\right]}$ & $ 1.54_{\left[0.02\right]}$  & $ 0.84_{\left[0.00\right]}$ & $ 7.10_{\left[0.04\right]}$ & $ \textbf{30.77}_{\left[0.14\right]}$\\ 
    $\text{DialoGPS}_{K=4}$ & $ \textbf{19.05}_{\left[0.18\right]}$ & $ \textbf{7.70}_{\left[0.16\right]}$ & $ \textbf{3.41}_{\left[0.09\right]}$ & $ \textbf{1.61}_{\left[0.07\right]}$  & $
    0.91_{\left[0.01\right]}$ & $ 7.45_{\left[0.09\right]}$ & $ 30.29_{\left[0.12\right]}$ \\
    $\text{DialoGPS}_{K=8}$ & $ 19.04_{\left[0.08\right]}$ & $ 7.64_{\left[0.11\right]}$ & $ 3.40_{\left[0.10\right]}$ & $ 1.60_{\left[0.08\right]}$  & $ 0.93_{\left[0.01\right]}$ & $ 7.64_{\left[0.06\right]}$ & $ 30.39_{\left[0.14\right]}$ \\\midrule\multicolumn{8}{c}{Multi-reference DailyDialog Dataset}\\\midrule
    Transformer & $33.93_{\left[0.26\right]}$ & $12.32_{\left[0.25\right]}$ & $4.93_{\left[0.23\right]}$ & $2.14_{\left[0.14\right]}$ & $2.59_{\left[0.03\right]}$ & $20.62_{\left[0.12\right]}$ & $ 35.79_{\left[0.15\right]}$\\
    ResBag  & $34.10_{\left[0.27\right]}$ & $12.61_{\left[0.18\right]}$ & $4.82_{\left[0.17\right]}$ & $2.13_{\left[0.13\right]}$ & $2.98_{\left[0.06\right]}$ & $24.44_{\left[0.17\right]}$ & $35.22_{\left[0.15\right]}$\\
    TSA   &$36.14_{\left[0.11\right]}$ & $13.21_{\left[0.15\right]}$ & $5.43_{\left[0.14\right]}$ & $2.46_{\left[0.13\right]}$ & $3.56_{\left[0.04\right]}$ & $26.89_{\left[0.21\right]}$ & $ 35.37_{\left[0.13\right]}$\\
    DD++  &$36.87_{\left[0.32\right]}$& $14.09_{\left[0.24\right]}$ & $6.13_{\left[0.23\right]}$ & $2.91_{\left[0.17\right]}$ & $3.84_{\left[0.03\right]}$ & $28.58_{\left[0.38\right]}$ & $\underline{37.04}_{\left[0.14\right]}$\\
    M\&D-D & $36.97_{\left[0.12\right]}$ & $14.28_{\left[0.09\right]}$ & $6.50_{\left[0.19\right]}$ & $3.28_{\left[0.17\right]}$ & $3.65_{\left[0.03\right]}$ & $25.35_{\left[0.21\right]}$ & $36.02_{\left[0.15\right]}$\\\midrule
    $\text{DialoGPS}_{K=1}$  & $37.21_{\left[0.12\right]}$ & $14.72_{\left[0.14\right]}$ & $6.65_{\left[0.12\right]}$ & $3.29_{\left[0.11\right]}$  & $4.25_{\left[0.05\right]}$ & $28.39_{\left[0.14\right]}$ & $36.14_{\left[0.08\right]}$\\
    $\text{DialoGPS}_{K=2}$ & $38.01_{\left[0.13\right]}$ & $14.79_{\left[0.07\right]}$ & $6.52_{\left[0.06\right]}$ & $3.20_{\left[0.04\right]}$  & $4.34_{\left[0.06\right]}$ & $29.04_{\left[0.25\right]}$ & $36.15_{\left[0.16\right]}$ \\
    $\text{DialoGPS}_{K=4}$ & $ 38.27_{\left[0.20\right]}$ & $ 14.77_{\left[0.13\right]}$ & $ 6.62_{\left[0.15\right]}$ & $ \textbf{3.33}_{\left[0.20\right]}$ & $ \textbf{4.53}_{\left[0.07\right]}$ & $ \textbf{30.18}_{\left[0.17\right]}$ & $36.09_{\left[0.08\right]}$\\
    $\text{DialoGPS}_{K=8}$ & $ \textbf{38.46}_{\left[0.18\right]}$ & $ \textbf{15.05}_{\left[0.23\right]}$ & $ \textbf{6.70}_{\left[0.24\right]}$ & $ 3.30_{\left[0.14\right]}$  & $ 4.32_{\left[0.06\right]}$ & $ 28.35_{\left[0.14\right]}$  & $35.82_{\left[0.16\right]}$\\
    $\text{DialoGPS}_{K=16}$ & $ 38.38_{\left[0.14\right]}$ & $14.89_{\left[0.06\right]}$ & $ 6.62_{\left[0.13\right]}$ & $ 3.30_{\left[0.15\right]}$  & $ 4.41_{\left[0.05\right]}$ & $ 29.84_{\left[0.08\right]}$ & $ 35.81_{\left[0.05\right]}$ \\\midrule\multicolumn{8}{c}{ Component Ablation on Multi-reference DailyDialog (K=4)}\\\midrule
    \textendash M.E. & $38.04_{\left[0.17\right]}$ & $15.00_{\left[0.12\right]}$ & $6.63_{\left[0.12\right]}$ & $3.21_{\left[0.11\right]}$ & $4.22_{\left[0.03\right]}$ & $28.05_{\left[0.10\right]}$ & $35.96_{\left[0.09\right]}$ \\
    \textendash M.D. & $34.62_{\left[0.12\right]}$ & $12.71_{\left[0.13\right]}$ & $5.20_{\left[0.08\right]}$ & $2.33_{\left[0.08\right]}$ & $3.19_{\left[0.04\right]}$ & $24.65_{\left[0.16\right]}$ & $35.14_{\left[0.13\right]}$\\
    \textendash Brown. & $38.05_{\left[0.22\right]}$ & $14.68_{\left[0.05\right]}$ & $6.36_{\left[0.04\right]}$ & $3.01_{\left[0.10\right]}$ & $4.05_{\left[0.09\right]}$ & $27.58_{\left[0.18\right]}$ & $35.52_{\left[0.11\right]}$\\
    \textendash M.E. \textendash Brown. & $38.42_{\left[0.13\right]}$ & $14.76_{\left[0.15\right]}$ & $6.55_{\left[0.05\right]}$ & $3.17_{\left[0.12\right]}$ & $4.11_{\left[0.03\right]}$ & $27.64_{\left[0.16\right]}$ & $36.12_{\left[0.12\right]}$\\
    \textendash M.D. \textendash Brown.  & $34.49_{\left[0.31\right]}$ & $12.68_{\left[0.28\right]}$ & $5.15_{\left[0.23\right]}$ & $2.29_{\left[0.17\right]}$ & $2.97_{\left[0.45\right]}$ & $24.46_{\left[0.15\right]}$ & $35.11_{\left[0.12\right]}$\\
    \textendash M.E. \textendash M.D. & $33.93_{\left[0.26\right]}$ & $12.32_{\left[0.25\right]}$ & $4.93_{\left[0.23\right]}$ & $2.14_{\left[0.14\right]}$ & $2.59_{\left[0.03\right]}$ & $20.62_{\left[0.12\right]}$ & $35.79_{\left[0.15\right]}$\\\bottomrule
    \end{tabular}}

    \caption{Automatic evaluation and ablation results on multi-reference DailyDialog and PersonaChat. We apply Top-5 Sampling decoding scheme. The standard deviation [$\sigma$] (across 5 runs) is also reported. In the ablation results table, M.E/D. stands for applying mixup in the encoder/decoder, and Brown. stands for optimizing $f_\theta$ with Eq.~\ref{eq:brownian-loss}. When there is no mixup in either encoder or decoder, the model degenerates into a vanilla transformer.}
    \label{tab:main}
\end{table*}

\section{Experimental Settings}
\paragraph{Datasets}

We conduct multi-turn dialogue generation experiments on two public datasets: DailyDialog~\citep{li-etal-2017-dailydialog} and PersonaChat~\citep{zhang-etal-2018-personalizing}. DailyDialog contains high-quality multi-turn dialogues collected from daily conversations, and it has many multi-reference versions~\cite{sai-etal-2020-improving, gupta-etal-2019-investigating} denoted by humans, which makes it possible for us to compare DialoGPS with human annotators. Besides, it is more reliable to evaluate the generalization and performance with multiple references. PersonaChat collects dialogues based on chatters' profiles. Profiles are not shown to models, so it is more challenging and open to generate proper responses, measuring generalization capacity better.

\paragraph{Baselines and Parameters} 
We compare DialoGPS with (1) Transformer~\cite{vaswani2017attention}. (2)DD++~\cite{sai-etal-2020-improving}: it is a  variant of DailyDialog in which each context has five manually denoted responses. We train a vanilla Transformer on it. (3) TSA~\cite{xie2022targetside}: it is an unsupervised augmentation method in the decoder side. It uses its decoder's output to construct pseudo-target data which is used to train the model for another round. From a dialogue generation viewpoint, it is a one-to-many method that bootstraps based on one-to-one data. (4) M$\&$D-D~\cite{zhang-etal-2020-dialogue}: it uses a pre-trained model and BM-25 algorithm to construct new context-response pairs from unpaired sentences. Since it is a single-turn augmentation, given a multi-turn dialogue, we only apply this method to the last two turns. (5) ResBag~\cite{qiu-etal-2019-training}: an augmented VAE-based model. It captures the common feature in the bag of plausible responses and then adds the specific feature to obtain the final output, which utilizes the multiple references better.

Because DialoGPS is a plug-and-play method, we add it to a $\text{BART}_{\text{Large}}$~\cite{lewis-etal-2020-bart} and compare with $\text{DialoFlow}_{\text{Large}}$~\cite{li-etal-2021-conversations}. DialoFlow is one of the state-of-the-art pre-trained models in open-domain dialogue generation. It augments the model by modeling the dialogue flow. More details on the implementation and hyper-parameters are in Appendix~\ref{apx:hyper}.

\paragraph{Evaluation Metrics}
We consider three automatic evaluation metrics: BLEU~\cite{papineni-etal-2002-bleu}, Distinct (DIST)~\cite{li-etal-2016-diversity}, and BLEURT~\citep{sellam-etal-2020-bleurt}. BLEU measures the word overlap between generated responses and the ground truth. DIST measures the ratio of unique n-grams in the generated responses. Because these two metrics are only sensitive to lexical variation, we evaluate BLEURT, an advanced learned semantic-sensitive evaluation metric based on BERT~\cite{devlin-etal-2019-bert}. On the evaluation of fine-tuning pre-trained models, we follow~\cite{li-etal-2021-conversations} to report METEOR~\cite{lavie-agarwal-2007-meteor} and Entropy~\cite{NEURIPS2018_23ce1851}.
For human evaluation, we recruit five evaluators to manually judge 200 samples from each experiment in blind testing, where we set three metrics to comprehensively evaluate the generation quality: whether a response is \textit{readable} (\textbf{Read.}), \textit{coherent} (\textbf{Coh.}), and \textit{informative} (\textbf{Info.}). For each aspect, evaluators can score at `bad', `borderline' and `good'.

\begin{table*}[t]
    \centering
    \vskip 0.15in
    \small
\begin{tabular}{@{}lcccccccccc@{}}
\toprule
\multirow{2}{*}{\textbf{Models}}& \multicolumn{4}{c}{\textbf{DailyDialog}} & \multicolumn{4}{c}{\textbf{PersonaChat}}\\
\cmidrule(l{2pt}r{2pt}){2-5}\cmidrule(l{2pt}r{2pt}){6-9}
     & \textbf{BLEU-2} & \textbf{BLEU-4}  & \textbf{METEOR}  &\textbf{Entropy} & \textbf{BLEU-2} & \textbf{BLEU-4}    & \textbf{METEOR}  &\textbf{Entropy} \\ \midrule
   BART & 27.87 & 10.85 & 14.69 & 9.29 & 9.95 & 3.38 & 8.69 & 6.55 \\
    DialoFlow & 28.02 & 11.57 & \textbf{16.40} & 9.46 
 & 10.46 & 3.03 & \textbf{9.32} & \textbf{6.89} \\\midrule
    \textbf{BART + DialoGPS} & \textbf{29.18} &\textbf{12.05} & 15.30 & \textbf{9.73} & \textbf{10.97} & \textbf{4.08} & 9.26 & 6.70\\\bottomrule
    \end{tabular}
    \caption{Automatic evaluation results on fine-tuning pre-trained models (beam search with width 5).}
    \label{tab:pretrain}
\end{table*}

\begin{table}[t]
    \centering
    \vskip 0.15in
	\small
\resizebox{\linewidth}{19mm}{
\begin{tabular}{@{}lcccccc@{}}
\toprule
\multirow{2}{*}{\textbf{Models}}& \multicolumn{3}{c}{\textbf{DailyDialog}} & \multicolumn{3}{c}{\textbf{PersonaChat}}\\
\cmidrule(l{2pt}r{2pt}){2-4}\cmidrule(l{2pt}r{2pt}){5-7}
      & \textbf{Read.} & \textbf{Coh.} & \textbf{Info.}   
     & \textbf{Read.} & \textbf{Coh.} & \textbf{Info.} \\ \midrule
    Transformer &70/8 &69/9 &73/12 & 53/14 & 51/11 & 52/9\\
    ResBag &58/13 &60/11 &64/14 &51/14 &50/19 & 51/16\\
    TSA   &59/15 &57/16 &60/16 &48/20 & 47/22 &43/20 \\
    DD++  &53/24 &55/20 &51/17  &-&-&-\\
    M\&D-D  &56/19 &47/20 &52/16 & 44/21 & 46/18 & 45/17\\\midrule
BART &40/34 &42/23 &44/26 &39/31 &41/26 &34/20 \\
DialoFlow &36/32 &40/29 &43/27 & 39/34 &35/28 &35/25 \\\bottomrule
    \end{tabular}}
    \caption{Human evaluation results (rounded). Compared with each baseline, we report our win/lose percentage. Evaluators achieve substantial agreement with kappa value $0.62$ on experiments trained from scratch and $0.70$ on pre-trained experiments.
    }
    \label{tab:human}
\end{table}

\section{Results}
Table~\ref{tab:main} shows the automatic evaluation results. On PersonaChat, without access to chatters' profiles, conversations are so open that there is so much noise in data for models to learn. Therefore, models prefer safe responses and thus DISTs are relatively low. However, DialoGPS still improves by about 20$\%$ in DISTs than the best-performing baseline. Also, BLEU and BLEURT scores imply that DialoGPS matches references more lexically and more semantically. 
On the multi-reference DailyDialog dataset, DialoGPS gains improvement by a large margin than other strong baselines. Also, most baselines suffer a trade-off between matching the references and diversifying responses. By contrast, DialoGPS performs evenly well on all metrics. DialoGPS also wins 6 out of all 7 metrics compared with the model trained on DD++, the human-written multi-reference training set. Our results in bold pass the significance test p $<$ 0.01. In Table~\ref{tab:pretrain}, when adding $\text{DialoGPS}_{K=2}$ to a pre-trained BART and fine-tuning on two datasets, it achieves competitive performance as one of the SOTA dialogue generation pre-trained models, DialoFlow. DialoFlow augments the generation with the help of `flow', i.e., the difference of adjacent utterances in continuous space. Their flows are not as flexible as paths sampled from the Brownian Bridge, which is one of the reasons that DialoGPS outperforms DialoFlow in five out of all eight metrics. Table~\ref{tab:human} shows human evaluation results. In three metrics, DialoGPS achieves the top rank with solid agreement among evaluators. More evaluation details are in Appendix~\ref{apx:eval}.

\subsection{Study on Dialogue Paths}
\begin{figure}[!t]
\centering
    \includegraphics[width=\linewidth]{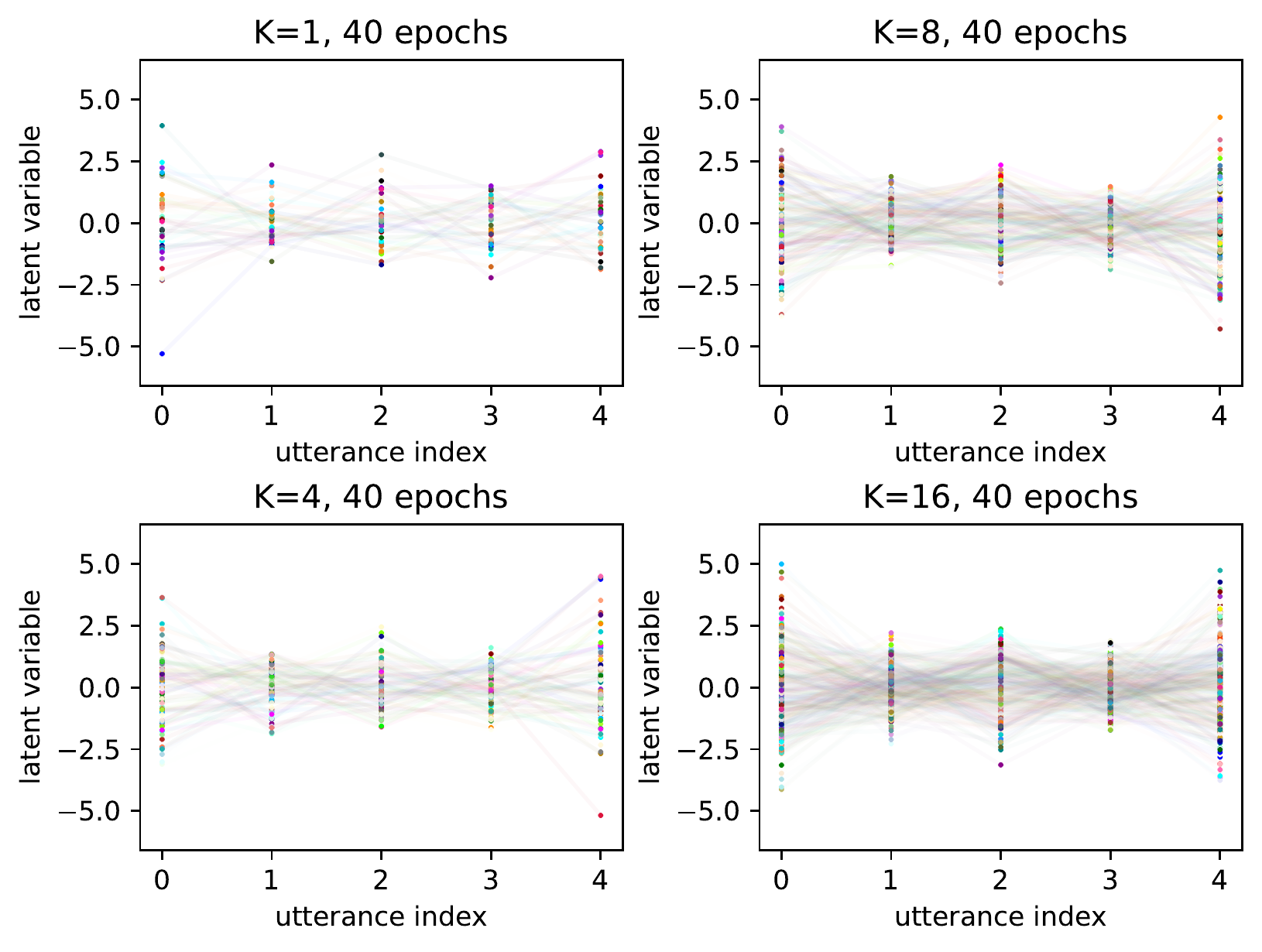}
\caption{The visualization of sampled dialogue paths (normalized expectations) for a 5-utterance dialogue, training with varying $K$.}
\label{fig:k}
\end{figure}

We conduct an ablation study on the number of sampled dialogue paths $K$,  results are shown in Table~\ref{tab:main}. On both datasets, with the increase of K, various metrics increase and then reach the bottleneck or slightly decrease. This phenomenon mainly dues to that different from discrete data, sampled paths in continuous space have a information bottleneck, i.e., if $K$ is big enough to cover the most samplable area in the Brownian Bridge, then increasing $K$ further may cause little improvement or even decrease due to more noise. We visualize the sampled paths of a conversation with 5 utterances during training in Figure~\ref{fig:k}. A sample at each time step is denoted as a point and paths are depicted. We can see that the Brownian Bridge area covered by paths is significantly increased when K increases from 1 to 8, but there is a slight difference when K further increases to 16. The visualization confirms automatic evaluation results in Table~\ref{tab:main}.

\subsection{Component Ablation}
We study the effect on the performance of the following components in DialoGPS: mixup in the encoder (M.E.), mixup in the decoder (M.D.), and constraints from Eq. \ref{eq:brownian-loss} that is the optimization of the mapping function (Brown.). The results are reported at the bottom of Table \ref{tab:main}. Removing mixup in the decoder (\textendash M.D.) degenerates DialoGPS to a many-to-one mode and thus the performance degrades much, confirming the intuition mentioned in $\S$\ref{sec:intro}. Removing mixup in the encoder(\textendash M.E.) degenerates DialoGPS to a one-to-many pattern which is insufficient compared with the many-to-many pattern, and DIST drops while the BLEU maintains. Nonetheless, the performance is still competitive with the best one-to-many baseline. Without constraints from Eq. \ref{eq:brownian-loss} (\textendash Brown.), there is no context-wise correlation among sampled latent variables and the mixup turns to introduce noise. This variant resembles sampling each utterance with a VAE~\cite{bowman-etal-2016-generating, miao2016neural}. However, Eq. \ref{eq:inferred-z} does not hold anymore so there exist gaps between the inference and the training, and results drop compared to the variant with Eq. \ref{eq:brownian-loss}. Overall, this variant still plays a positive role because adding noise during training is proved to be effective in improving the robustness and generalization of the model~\cite{JMLR:v15:srivastava14a, gao-etal-2021-simcse}. When there is neither M.D. nor M.E., the method becomes a vanilla transformer.

\subsection{Study on Utterance Representation}
\label{sec:utterance-representation}

\begin{table}[!t]
\centering
\small
\begin{tabular}{lcccc}
\toprule
\textbf{Method}
      & \textbf{BLEU-2} & \textbf{BLEU-4}   & \textbf{DIST-1} & \textbf{DIST-2} \\ \midrule
    Avg. & 14.77 & 3.33 & 4.53 & 30.18\\
    Avg. + Pos. & 14.41 & 2.89 & 4.19 & 29.22\\
    GPT-2  & 15.13 & 3.28 & 4.23 & 29.55\\
\bottomrule
    \end{tabular}
    \caption{Experimental results with different utterance representation methods (K=4).}
    \label{tab:sen}
\end{table}

In $\S$\ref{sec:mixup}, we defer details on obtaining utterance representations of each turn in a dialogue. We study three variants of encoding an utterance: (1) average embeddings of each token in an utterance (Avg.), (2) average embeddings of each token in an utterance along with position embeddings (Avg. + Pos.), and (3) encode utterances by a GPT-2~\cite{radford2019language}. We conduct this study on the multi-reference DailyDialog dataset and the results are in Table~\ref{tab:sen}. The simplest method (Avg.) achieves first place. With extra positional information, the performance drops a little, and in this experiment, we observed that the $\mathcal{L}_{\beta}$ term in the overall training objective Eq.~\ref{eq:obj} maintains steadily, but other terms increase a little. An explanation is that features to be mixed with latent variables ($e$ and $d$) have included positional information and positional information in latent variables introduces redundancy. For (GPT-2), we add a special token `<eou>' at the end of an utterance and view its corresponding output as the utterance representation. (GPT-2) costs much more training time and only beat (Avg.) in one metric. We guess there is an expression capacity gap so we try to (1) train a 4-layer language model to replace the GPT-2 and (2) apply GPT-2 in pre-trained experiments. In both experiments, we do not observe improvement than (Avg.). To sum up, the simplest (Avg.) achieves the best trade-off between performance and costs so in DialoGPS, we adopt this scheme by default. 

\subsection{What Does the Model Learn from Augmented Data?}
\label{sec:sample-infer}

If we mixup with sampled variables instead of expectations during inference, the model obtains the ability to generate diverse responses. Although we do not know what discrete labels augmented data have, to some extent the diverse outputs during inference reflect semantics that augmented data have during training. We provide a case in Table~\ref{tab:case}. Transformer and ResBag generates incoherent responses, and TSA answers the arrival time but not the way. DD++ reply to the context but does not leads to the follow-up dialogue. M$\&$D-D responds properly but can only provide one answer. 
We let DialoGPS generate 10 times and report all the outputs along with their respective frequency.

\begin{table}[t]
    \centering
    \vskip 0.15in
    \resizebox{\linewidth}{29mm}{
    \begin{tabular}{|c|l|}
   \hline
      \multirow{4}{*}{\diagbox{$x_3$}{$X_{0:2}$}} &A: Excuse me, sir. Is there a barber near here?\\
       &B: Yes, the nearest one is at the third cross of this road.\\
      & A: I'm a stranger here. How can I get there, please?\\
        & B: \_\_\_\_\_\_\_\_\_\_\_\_\_\_\_\_\_\_\_\_\_\_\_\_\\\hline
                Transformer & Thank you very much.\\ \hline
        ResBag & Two stops at the next door.\\ \hline
        TSA & Let me see. It's about ten minutes.\\ \hline
        DD++ & Sure.\\\hline
        M\&D-D & You can take the subway to get there.\\\hline
        \multirow{7}{*}{DialoGPS} & You have to go to the next stop. ($\times$2)\\ 
        & You get off at the next stop. ($\times$2)\\
        & You have to change. ($\times$2)\\ 
        & You have to go to the hotel. ($\times$1)\\ 
        & It's not easy. You have to go. ($\times$1)\\ 
        & You have to go to the airport. ($\times$1)\\ 
        & Then, you have to go to the hotel. ($\times$1)\\ \hline
    \end{tabular}}
    \caption{10 outputs given by DialoGPS when adopting sampling then mixup during inference. To avoid the randomness introduced by the decoding strategy, responses are decoded by Beam Search with width 5.}
    \label{tab:case}
\end{table}

The frequency, the semantics, and lexical features of responses resemble a Gaussian distribution. In this case, `you have to go to (get off at) the next stop' is close to the expectation. As the semantics get farther away, the frequency of other responses are lower.
Overall, DialoGPS provides diverse choices to arrive at the barber. This case shows that continuous augmented data do have open dialogue knowledge which is conducive to model generalization. 

\section{Conclusion}
We propose DialoGPS that first augments open-domain and multi-turn dialogue generation from a many-to-many perspective. Specifically, We map dialogues into the continuous semantic space which is modeled by our extended Brownian Bridge and sample dialogue paths to augment training. We propose a self-distillation framework to utilize augmented data despite the inaccessible discrete labels. Empirically, we prove the effect of DialoGPS and study its characteristics. DialoGPS could be a general method that suits seq2seq tasks where the source has multiple sentences and the target is different from the source in semantics, like summarization. However, DialoGPS should be modified according to the unique properties of the task, which is left to study in the future. 

\section*{Limitations}
Similar to other augmentation methods, DialoGPS demands high requirements for computing resources. The training is performed on up to 8 V100 GPUs. On DailyDialog: a vanilla transformer only needs 50 minutes while a non-pretrained DialoGPS takes about 80 minutes when $K=1$. Other baselines take about the same amount of time as DialoGPS $K=1$. But when DialoGPS achieves its performance peak ($K=16$), the training takes 4 hours. Most of time cost comes from sampling which is difficult to be accelerated by GPUs. 

\section*{Acknowledgement}
This work was supported by National Natural Science Foundation of China (NSFC Grant No. 62122089), Beijing Outstanding Young Scientist Program NO. BJJWZYJH012019100020098, and Intelligent Social Governance Platform, Major Innovation $\&$ Planning Inter-disciplinary Platform for the "Double-First Class" Initiative, Renmin University of China.

\bibliography{custom}
\bibliographystyle{acl_natbib}

\appendix
% \clearpage
\newpage
\section{Appendix}
\subsection{Model Implements}
\label{apx:hyper}
In pre-process, we truncate the original long conversations in the dataset with the window size 5. Table~\ref{tab:datasets} shows the dataset statistics.

\begin{table}[!htbp]
\centering
\small
\begin{tabular}{@{}lccc@{}}
\toprule
\textbf{Datasets} & \multicolumn{1}{c}{\textbf{Train}} & \multicolumn{1}{c}{\textbf{Valid}} & \multicolumn{1}{c}{\textbf{Test}}\\ 
\midrule
        DailyDialog & 44050 & 4176 & 6740(Multi-ref) \\
        PersonaChat & 68859 & 8593 & 8239 \\ \bottomrule
\end{tabular}
\caption{Dataset statistics.}
\label{tab:datasets}
\end{table}

For non-pretrained experiments, our code is based on fairseq~\cite{ott2019fairseq}. We adopt grid search to tune hyper-parameters. On the DailyDialog dataset, the search ranges for learning rate and batch size are $\{0.00008,0.00010, 0.00012,0.00015\}$ and $\{112, 160\}$, respectively. On the PersonaChat dataset, the search ranges for learning rate and batch size are $\{0.00010, 0.00012,0.00015\}$ and $\{32, 64\}$, respectively. We choose the parameter combination with the lowest perplexity in the validation set. Table~\ref{apxtab:hyper} shows the searched results for each experiment.

\begin{table}[ht]
\centering
\small
\resizebox{\linewidth}{20mm}{
\begin{tabular}{@{}lcccc@{}}
\toprule
\textbf{Method}
      & \textbf{LR(DD)} & \textbf{Batch size(DD)} & \textbf{LR(PS)} & \textbf{Batch size(PS)}\\ \midrule
    Transformer & 1e-4 & 112 & 1e-4& 32\\
    ResBag & 8e-5 & 160 & 1e-4 & 64\\
    TSA & 8e-5 & 160 & 1.5e-4 & 32\\
    DD++& 8e-5 & 112 & - & -\\
    M$\&$D-D & 1e-4 & 112 & 1e-4 & 64\\
    $\text{DialoGPS}_{K=1}$ &1.5e-4 & 160 &1.5e-4 & 64\\
    $\text{DialoGPS}_{K=2}$ &1.5e-4 & 160 &1e-4 & 64\\
    $\text{DialoGPS}_{K=4}$ & 1.5e-4 & 112 &1.2e-4 & 64\\
    $\text{DialoGPS}_{K=8}$ & 1.5e-4 & 160 &1.2e-4 & 64\\
    $\text{DialoGPS}_{K=16}$ & 8e-5 & 160 & - & -\\\bottomrule
    \end{tabular}}
    \caption{Learning rate and batch size in each experiment.}
    \label{apxtab:hyper}
\end{table}

Except for batch size and learning rate, the following important settings: the warmup steps are 4000. We use Adam optimizer with $\beta = (0.9,0.98)$. Both attention dropout and activation dropout are 0.1. For models trained from scratch, $\delta$ on Dailydialog is $\frac{1}{2}$ and $\frac{1}{3}$ on PersonaChat. For fine-tuned models, $\delta$ is $\frac{1}{2}$ on two datasets. We select the best checkpoint based on the perplexity in the validation set. Early stop patience is 10 epochs. For pre-trained experiments, on both datasets, the batch size is 64 and learning rate is $0.00002$. The training is performed on Nvidia V100 GPU. On DailyDialog: our method takes about 80 minutes when $K=1$, 4 hours when $K=16$, and 8 hours to finetune a $\text{BART}_\text{large}$.

Because M$\&$D-D does not suit multi-turn settings, we only use it to modify the last two turns with Okapi BM25 algorithm and we finetune BERT on DailyDialog and PersonaChat respectively to measure the fluency between the last two utterances and the fluency between the penultimate sentence and the above as filtration. In our experiments, on two datasets, the paired sentence set $D_p$ is same as the original training set and the unpaired sentence set $D_u$ is constructed from all sentences in DD++. 
On DailyDialog, we use multiple references in DD++ as the response bag of ResBag, and on PersonaChat, we use constructed data from M$\&$D-D as its response bag. 
\begin{table}[t]
    \centering
    \small
    \begin{tabular}{@{}lcc@{}}
    \hline
      \textbf{Method}   &  \textbf{PersonaChat} & \textbf{DailyDialog}\\\hline
     Transformer  & 2.93 & 3.08\\
     ResBag & 2.93 & 3.12 \\
     TSA & 2.92 & 3.13 \\
     DD++ & - & \textbf{3.24} \\
     M$\&$D-D & 2.96 & 3.13 \\\hline
     \textbf{DialoGPS(K=4)} & \textbf{3.03} & \textbf{3.24} \\\hline
    \end{tabular}
    \caption{QuantiDCE results on two datasets.}
    \label{tab:my_label}
\end{table}

\subsection{Evaluation Details}
\label{apx:eval}

Because some evaluation script links of DialoFlow~\cite{li-etal-2021-conversations} are out of date, we can not reproduce NIST~\cite{lin-och-2004-automatic} scores so we do not report it. This issue was also reported by the community~\footnote{\url{https://github.com/microsoft/DialoGPT/issues/72}}. Also, METEOR and Entropy are reproduced. Our reproduced BLEU scores are close to the original paper so we directly quote their results.

Our human evaluators are recruited from Amazon Mturk. In terms of human evaluation, all generated responses are re-capitalized and de-tokenized fairly. The salary for each evaluator is 1 dollar per 10 samples. To give a fair salary, we first evaluate 50 samples by ourselves, calculate the time and effort, and set this amount (samples evaluated by ourselves are just for evaluating the salary, which is not given to evaluators and not reported in the final results).

\subsection{QuantiDCE}
\label{apx:quantidce}
In addition to the metrics mentioned in the main paper, we further supplement our evaluation with the dialogue-specific metric QuantiDCE~\cite{DBLP:journals/corr/abs-2106-00507}, which measures the coherence between the response and the context. The results show that our proposed DialoGPS outperforms all baseline models.

\end{document}